# Navigating Uncertainties with Probabilistic Diffusion-Based Motion Prediction and Active Inference

Yufei Huang, Yulin Li, Andrea Matta and Mohsen Jafari

*Abstract*— **This paper presents a novel approach to improving autonomous vehicle control in environments lacking clear road markings by integrating a diffusion-based motion predictor within an Active Inference Framework (AIF). Using a simulated parking lot environment as a parallel to unmarked roads, we develop and test our model to predict and guide vehicle movements effectively. The diffusion-based motion predictor forecasts vehicle actions by leveraging probabilistic dynamics, while AIF aids in decision-making under uncertainty. Unlike traditional methods such as Model Predictive Control (MPC) and Reinforcement Learning (RL), our approach reduces computational demands and requires less extensive training, enhancing navigation safety and efficiency. Our results demonstrate the model's capability to navigate complex scenarios, marking significant progress in autonomous driving technology.**

## I. INTRODUCTION

In tackling the challenge of autonomous navigation under uncertain and opposing circumstances, our research adopts a novel approach by utilizing new advances in Generative AI, namely Probabilistic Diffusion (PD), and Active Inference (AIF). PD reverse engineers a motion predictor and AIF safely guides vehicles to their intended destinations. We conjecture that the proposed approach can be generically applied to many engineering applications involving predictions and control. In this article, however, we demonstrate the main ideas for vehicle navigation in a parking lot setting, where vehicles are expected to park at some designated spots. Lacking clear navigational cues and given random interactions between vehicles, parking lots can serve as a building block for mirroring some of the complexities of autonomous navigation in more complex settings, such as unmarked roadways. We propose basic ideas for extending our parking lot model to the unmarked roadway scenarios but leave the details to a future manuscript. Our approach is pioneering, particularly in the utilization of PD for its novel application in the realm of autonomous navigation, beyond its traditional use in image processing where it excels in generative AI tasks and reverse engineering complex systems [1]. Notably, this research marks one of the first instances of PD being adapted for reverse engineering applications, a testament to its versatility and robust generative capabilities [2]. Similarly, AIF is employed not just as a control strategy but as a cognitive model that mimics the Predictive Mind, enhancing decision-making under uncertainty through predictive reasoning [3].

Uncertainties, such as missing roadway markings and mixed traffic, set a higher demand for a robust perception model in autonomous navigation. Human drivers utilize their inner predictive mind [4] capability to predict and minimize consequential errors by properly acting according to their perception of the roadway conditions. By anticipating and predictive reasoning, human drivers can handle poor road conditions and avoid random moving traffic and parked vehicles.

Navigating through traffic safely and efficiently remains a paramount concern for autonomous navigation, especially in environments where traditional road markings are not clear. This challenge becomes even more pronounced in mixed flow traffic environments, where autonomous vehicles must coexist with human-driven vehicles, all navigating without the guidance of clear lane markings. Traditional navigation systems, which rely heavily on well-defined road infrastructures, often fall short under these conditions. The problem at hand focuses on enabling autonomous vehicles to find safe and efficient paths to their destinations in such unmarked road segments, hence, an ability to adapt to less structured environments.

Our approach sets itself apart from traditional methods such as Model Predictive Control (MPC) and Reinforcement Learning (RL), by leveraging the strengths of diffusion models and the AIF. MPC relies heavily on precise vehicle modeling and the resolution of complex optimization challenges. Unlike RL, which necessitates extensive training, our model offers a direct, structured method for predicting vehicle trajectories, incorporating safety considerations through its handling of predictive uncertainty. This unique combination of diffusion models and AIF, with its ability to make informed decisions under uncertain conditions, positions our model as a pioneering (the first of its kind) solution in the realm of autonomous navigation.

This paper is structured as follows: Section II reviews related works in the field; Section III details the methodology behind our diffusion-based motion predictor and the AIF controller; Section IV presents the results from our simulation studies, and Section V concludes the paper with a discussion on the implications of our findings and directions for future research.

## II. RELATED WORKS

Y. Huang, Y. Li, and M. Jafari are with the Department of Industrial and Systems Engineering, Rutgers University – New Brunswick, Piscataway, NJ 08854, USA (e-mails: yh639@scarletmail.rutgers.edu; yl959@soe.rutgers.edu; jafari@soe.rutgers.edu).

A. Matta is with the Department of Mechanical Engineering of Politecnico di Milano – Via La Masa 1 20156, Milano, ITALY (e-mail: andrea.matta@polimi.it).

Navigating unmarked road environments poses unique challenges for autonomous vehicles, as traditional cues used for lane following and distance keeping are not available. Model Predictive Control (MPC) has been extensively applied in autonomous vehicle navigation due to its ability to handle dynamic constraints and predict future vehicle states [5]. For instance, [6] demonstrated MPC's efficacy in lane-keeping and obstacle avoidance by incorporating real-time traffic data into the control strategy. However, MPC's performance heavily relies on the accuracy of the vehicle model and the computational complexity of solving optimization problems in real-time [7]. In scenarios with undefined road markings, the absence of structured environmental data can limit MPC's predictive accuracy, making it less adaptable to unforeseen changes in traffic flow or road conditions. Recent studies primarily focus on the development of robust machine learning models that can interpret complex environments where traditional sensor-based systems falter. Reinforcement Learning (RL) has been praised for its adaptability and ability to improve over time, as highlighted by [8], who successfully applied deep reinforcement learning for trajectory planning in automated parking systems. But RL requires extensive training data and significant computational resources, especially in complex and dynamic environments.

Diffusion models have been predominantly utilized in image processing and generation fields, as detailed by [9]. Our work extends the application of diffusion models to the domain of reverse engineering; The diffusion model is used as a generative model for next state prediction in AIF. AIF integrates perception, action, and cognition into a cohesive framework, emphasizing the role of uncertainty and the agent's internal model in guiding its behavior. The AIF has been used in robotics and cognitive science to model decision-making processes under uncertainty [10]. Unlike Reinforcement Learning, which aims to maximize a numerical reward signal through actions, AIF takes actions as a means to minimize the expected free energy. This fundamental difference shifts the focus from seeking rewards to reducing uncertainty and achieving a state of least surprise [11]. In RL, decisions are driven by the potential for reward maximization, often defined in terms of explicit rewards linked to specific outcomes. However, AIF embeds a preference-based approach where no explicit reward signal is necessary; instead, it operates under a model where rewards are integrated as preferences over sensory states, known as free energy. By minimizing free energy, AIF inherently balances exploration and exploitation [12], adapting its strategy based on both current understanding and new observations. This holistic approach allows agents to not only respond to their environment but also anticipate changes, making decisions that are informed by both past experiences and potential future states.

While AIF's application to autonomous vehicles is nascent, preliminary studies, such as those by [13] [14], indicate its promise for enhancing adaptive decision-making in dynamic environments. Our research contributes to this emerging field by integrating AIF with a diffusion-based motion predictor for improved navigation in mixed flow traffic environments. While existing literature provides a foundation for autonomous vehicle navigation, there is a distinct lack of research focused on mixed flow environments with unclear or no road markings. Furthermore, the potential of diffusion models and AIF in this context has not been fully explored, underscoring the novelty and significance of our approach.

## III. VEHICLE CONTROL IN UNMARKED PARKING AREAS

The goal of the control framework is to replicate a road segment devoid of lane markings. To mimic the scenario, we break down the control system to aid vehicle navigation in a simulated parking lot environment, where vehicles need to drive in unmarked corridors and avoid collision with other parked and moving vehicles. The explanation is split into three main parts. To begin with, we elaborate on the transformation of a conventional road structure into a parking lot configuration. This step is crucial in simulating a scenario where vehicles, starting from random positions, velocities, and directions, need to navigate towards their designated parking spots. This scenario represents a transition from a state of chaos to one of order, which requires skillful maneuvering by the vehicles to avoid stationary and mobile obstacles. Next, we introduce an innovative diffusion-based motion predictor. This predictor is engineered to calculate the probability distribution of a vehicle's imminent actions that will lead to its successful parking. The model is developed using diffusion model methodologies, which include a forward training process and a reverse application process. This dual-phase approach ensures a robust predictive framework capable of accurately forecasting vehicular movement within the parking lot. Lastly, we elucidate the application of the diffusion model within an active inference framework. Here, the diffusion model's generative capabilities are used to help vehicles choose the most optimal action at each junction based on the principle of expected free energy. Additionally, the model is continuously refined via variational free energy adjustments, enhancing navigational efficacy. Finally, the detailed workflow of the diffusion-based active inference framework for autonomous vehicle navigation is illustrated.

### A. Adapting Parking Lot Dynamics to Simulate Unmarked Road Navigation

Within the context of autonomous navigation each vehicle must not only determine a path that avoids collisions but also progress toward its destination amongst a dynamic and unpredictable setting. To tackle this, the concept of discretization becomes valuable. If we imagine breaking down the continuous road into smaller, manageable pieces, akin to segments on a game board, the problem becomes less daunting. As is shown in Fig.1 (a) and (b), within each of these discrete segments, the vehicle's immediate task is to determine a safe and viable route to the edge of the segment. This step-by-step approach, where the road is segmented into pieces, lays the groundwork for drawing parallels with a parking lot scenario. In a parking lot, vehicles navigate through aisles to reach a specific parking spot without the guidance of painted lines. Each aisle can be thought of as a segment of the road. The vehicles must maneuver with care, negotiating their way around other cars and obstacles, all while making progress toward their allocated parking space. Both scenarios share fundamental similarities: they require the vehicles to create order from disorder, forming structured outcomes—whether it be a neatly parked car or a vehicle successfully reaching the end of a road segment—out of initially unstructured situations.

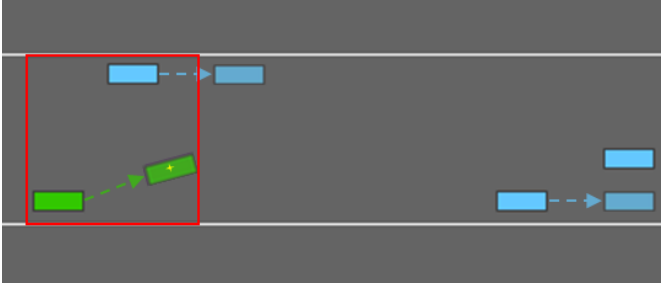

Fig. 1 (a) - Road segment for the controlled green car at timestep $n$

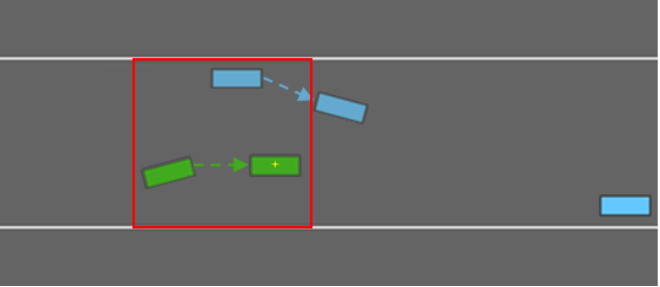

Fig. 1 (b) - Road segment for the controlled green car at timestep $n + 1$.

## B. Path prediction: the diffusion-based motion predictor

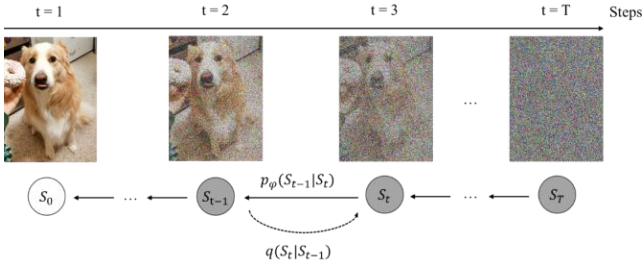

Fig. 2. Diffusion process for image generation.

The stable probabilistic diffusion [15] for image processing commonly contains two phases as is shown in Fig.3. The forward process begins with an original image, labeled as $X_0$, that is fully observable. Through a sequence of transformations, random perturbation (noise) is incrementally introduced to this image, leading to a progression of states where the original image becomes less recognizable, reaching a point of maximum randomness at $X_T$. Denote $\beta_t$ as the variance of the Gaussian noise to be added at timestep $t$, which is increasing over time, and the mean is defined as a scaled version of the previous state $x_{t-1}$, which is $\sqrt{1-\beta_t}x_{t-1}$. The probability distribution of the image at time $t$ can be represented as:

$$q(x_t|x_{t-1}) = \mathcal{N}(x_t; \sqrt{1-\beta_t}x_{t-1}, \beta_t I) \qquad (1)$$

In the subsequent reverse phase, the process methodically retracts the randomness, incrementally reinstating structure and clarity. While the graph concludes with an image that is partially clarified at $X_{t-1}$, the aim is to recover a clear image, closely resembling the initial state $X_0$. This process demonstrates the potential to reconstruct the original image from a state of maximal entropy through a systematic removal of the introduced noises. In practice, a neural network can be introduced to learn the probability distribution to iteratively denoise the image during the generative phase, with $\varphi$ representing the learned parameters of the model. The model for the reverse process can be represented as:

$$p_\varphi(x_{t-1}|x_t) = \mathcal{N}(x_{t-1}; \mu_\varphi(x_t,t), \sigma_\varphi^2(x_t,t)I) \qquad (2)$$

Drawing on this established framework, the development of the motion diffusion predictor adapts these principles to the realm of vehicular movement. Instead of transitioning between visual pixels, the motion diffusion predictor applies a similar iterative process to the kinematic variables of a vehicle during navigation.

### a. Forward process of the diffusion-based motion predictor

In contrast to the image generation approach, where clarity is progressively diminished by overlaying Gaussian noise, the motion diffusion predictor simulates the real-world conditions that a vehicle starts from its parking spot, note the state as $S_0$, and take random actions to drive away from the initial position. The final state $S_T$ is determined when a collision happens, whether the vehicle hit the boundary of the parking area, or a collision happen with other vehicles.

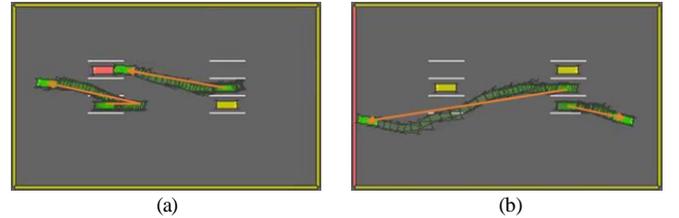

Fig. 3. Forward process for automated parking.

Illustrated in Fig.3 (a), the path of two actively moving vehicles is traced (indicated in green). Initially positioned in designated spots alongside four stationary vehicles, these two green vehicles perform random throttle and steering adjustments. Meanwhile, the yellow and red vehicles remain stationary. The depicted sequence concludes with one green vehicle making contact with the red vehicle. Fig.3 (b) provides an alternative depiction, where the roles are assumed by two yellow stationary cars and two green mobile cars. The green cars persist in random movements until an eventual collision with the parking area's boundary, represented by a red wall. This forward process effectively emulates the journey from a state of complete organization to one of disorder, analogous to the method by which noise is added in image processing.

Assuming that every aspect of the setting is visible and trackable, the model processes time in fixed, uniform segments. Within these segments, the system monitors all vehicles' conditions. Each vehicle's condition is characterized by its location, marked by coordinates $(x, y)$, its speed in the direction of both coordinates $(v_x, v_y)$, and the direction it's facing, noted as $h$. The state of each vehicle at a certain timestamp $t$ can be represented as a vector of 5 elements $S_t = [x_t \quad y_t \quad v_x^t \quad v_y^t \quad t_t]$. To aid in navigation, two extra pieces of information are provided for each vehicle at time $t$: $\theta_t$ shows the direction to the vehicle's starting parking spot, and $l_t$ measures how far the vehicle is from this spot.

During each discrete interval of time in the simulation, the vehicles under control undergo random changes in speed and direction. This is analogous to how, in the image diffusion process, the amount of Gaussian noise is increased over time to gradually obscure the image. In a similar vein, the range within which these random driving decisions are made becomes wider as time progresses. The sequence of random driving decisions made throughout this process is represented by $a_1, \cdots, a_{T-1}$. Each action $a_i$, where $i$ ranges from 1 to $T - 1$, is drawn from a Gaussian distribution that is truncated. To avoid dramatic movement change, the previous action is the mean of the current action, the actions are chosen based on the previous action $a_{i-1}$ but with added variability defined by $\sigma_i^2$. Mathematically, this is written as:

$$a_i \sim \mathcal{N}(a_{i-1}, \sigma_i^2 I) \tag{3}$$
$$a_i = clip(a_i, \ lb, \ ub) \tag{4}$$

The clip function applied here ensures that each action $a_i$ stays within a specific range, denoted by the lower and upper bounds ($lb$ and $ub$). This range reflects the real-world physical constraints on how much a vehicle can accelerate or decelerate (throttle) and turn (steering) at any given moment.

To manage the growing variability in the actions over time, a straightforward method increases $\sigma_i$ linearly from a starting low point to a peak. This increment allows the range of potential actions to widen as the simulation progresses, facilitating a gradual intensification of action diversity. The formula to calculate $\sigma_i$ reflects this linear growth, ensuring that with each step from the first to the last, the variance expands smoothly from its minimum to its maximum value:

$$\sigma_i = \sigma_{min} + (\sigma_{max} - \sigma_{min}) \times \frac{i}{T-1} \tag{5}$$

In Equation (5), $\sigma_{min}$ and $\sigma_{max}$ define the bounds of variance, while $i$ represents the current step, and $T - 1$ signifies the total number of steps. This approach guarantees a controlled and predictable escalation in action variability, mirroring the real-world scenario where decision-making might become increasingly bold or cautious as conditions evolve.

### b. Reverse process of the diffusion-based motion predictor

In the reverse process, akin to the denoising steps in traditional diffusion models, a vehicle starts at random positions inside the parking area with an initial speed and direction. The motion diffusion predictor employs learned parameters to infer the most probable previous action distribution of a vehicle - essentially 'denoising' the vehicle's trajectory to yield a predicted path back to its parking state. The predictor is trained on reversed state-action sequences:

$$S_T' \xrightarrow{a_{T-1}'} S_{T-1}' \xrightarrow{a_{T-2}'} \cdots \xrightarrow{a_1'} S_1' \xrightarrow{a_0'} S_0' \tag{6}$$

where $S_t'$ is the reverse of $S_t$ defined as:

$$S_t' = \begin{bmatrix} x_t & y_t & -v_t^x & -v_t^y & \pi - h_t \end{bmatrix} \tag{7}$$

Due to the dynamics of a vehicle's axles, the sequence of actions taken during the forward phase may differ from those in the reverse phase, implying that the reversed action $a_t' \neq a_t$. In the reverse process, the goal is to uncover the range of possible actions that would logically return the vehicle to its prior state $S_{t-1}'$ based on the current state and navigational aids $\theta_t$ and $l_t$. The aim here is mathematically modeled by seeking a distribution for actions that reconcile with the earlier state, given the existing conditions and guidance parameters. This approach endeavors to map out a backward trajectory, identifying actions that could have preceded the current vehicular state, hence facilitating a methodical backtracking to the initial position. The objective function, Kullback-Leibler (KL) Divergence [16], quantifies how one probability distribution diverges from a second, expected probability distribution. In this context, minimizing KL divergence helps in adjusting the parameters of the predictive model $Q$ so that it closely approximates the true distribution $P$, leading to more accurate predictions of the previous actions based on the given state and guidance features.

$$\min_{a_{t-1}'} KL(P(a_{t-1}'|S_t', \theta_t', l_t')||Q(a_{t-1}'|S_t', \theta_t', l_t')) \tag{8}$$

To effectively reduce the difference between what our model predicts and what actually happens, the goal is to closely match the state that our model predicts for the next step in the reverse process, $S_{t-1}'$ (which corresponds to the state just before the current state $S_t'$ in the forward process), with the actual previous state from the forward process. By doing so, we aim to refine our model's ability to accurately forecast the results of its suggested actions, ensuring the transitions it predicts align well with real-world transitions.

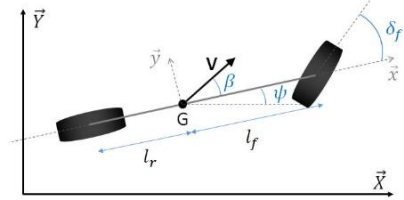

Fig. 4. Kinematic bicycle model.

Given the current state $S_t'$ and predicted action $a_{t-1}'$, the next state $S_{t-1}'$ can be estimated using the kinematic bicycle model [17] as shown in Fig.4. Define the vehicle's position as $(x, y)$, vehicle's forward speed as $v$, vehicle's heading as $\psi$, the vehicle's acceleration as $a$, vehicle's slip angle at the center of gravity $\beta$, and $\delta$ as the front wheel angle used as a steering command. The traditional bicycle model does not account for uncertainties in state transitions that occur due to factors such as slippery roads and tires. To address this, the model is enhanced by incorporating stochastic elements into its dynamic equations. Specifically, noise terms are introduced, assumed to follow a normal distribution, $\epsilon \sim \mathcal{N}(0, \Sigma)$, where $\Sigma$ represents a diagonal covariance matrix. The variances $\sigma_x^2$, $\sigma_y^2$, $\sigma_{v_x}^2$, $\sigma_{v_y}^2$, and $\sigma_\delta^2$ correspond to the respective state variables in $S_t'$. These modifications enable the model to generate the next state by integrating the impact of environmental and vehicular variabilities more realistically. The extended dynamic equations to get the next state can be written as:

$$x_{t+1} = x_t + v_{xt} \cdot cos(\delta_t + \beta_t) \cdot \Delta t + \epsilon_x \tag{9}$$
$$y_{t+1} = y_t + v_{yt} \cdot sin(\delta_t + \beta_t) \cdot \Delta t + \epsilon_y \tag{10}$$
$$\delta_{t+1} = \delta_t + \frac{v_t \cdot sin(\beta_t)}{L/2} \cdot \Delta t + \epsilon_\delta \tag{11}$$

$$v_{t+1}^x = v_t \cdot cos(\delta_{t+1}) + \epsilon_{v_x} \quad (12)$$

$$v_{t+1}^y = v_t \cdot sin(\delta_{t+1}) + \epsilon_{v_y} \quad (13)$$

where $\Delta t$ is the time step, $L$ is the length of the vehicle, $\beta_t = \arctan(\frac{1}{2} \cdot \tan(\delta_t))$ is the steering angle at the mass center, $v_t = \sqrt{v_{xt}^2 + v_{yt}^2}$ is the speed of the vehicle. $\epsilon_x, \epsilon_y, \epsilon_{v_x}, \epsilon_{v_y}$, and $\epsilon_\delta$ represents the noise on each state variable.

In the context of autonomous vehicle navigation using probabilistic methods, the application of Probabilistic Diffusion (PD) through score matching [18] offers an innovative way to handle the inherent randomness in vehicle dynamics. This approach utilizes a stochastic differential equation framework to model the dynamics of vehicle states, which is particularly useful in environments where precise control and prediction of vehicle behavior are critical due to unpredictable road conditions. The probabilistic state transition dynamics of a vehicle, represented by the bicycle model $S_t = F(S_{t-1}, A_{t-1})$

$$S(t + \Delta t) \cong S_t + f(S_t, A_t) \cdot dt + \sigma(A(t), t) \quad (14)$$
$$\cdot \sqrt{\Delta t} \mathcal{N}(A_{t-1}, \sigma_t^2 I, lb, ub)$$

$$dS_t = f(S_t, A_t) \cdot dt + g(A_t, t)d\widehat{\omega}_t \quad (15)$$

Here, $f(S_t, A_t)$ denotes the deterministic evolution of the state, reflecting predictable changes based on the current state $S_t$, action $A_t$, and the bicycle model $f(\cdot)$. The function $g(A_t, t)$ represents the diffusion term, introducing randomness into the process to account for environmental uncertainties and the inherent variability in vehicle responses. The term $d\widehat{\omega}_t$ denotes the increment of a Wiener process, encapsulating the random fluctuations that affect the vehicle's trajectory. To address the challenge of modeling the reverse process, where one aims to infer past states from current observations, a score matching technique is integrated into the dynamics:

$$dS_t = [f(S_t, A_t) - g(A_t, t)^2 \nabla_{S_t} \log q_t(S_t | S_0)] dt \quad (16)$$
$$+ g(A_t, t) d\widehat{\omega}_t$$

In this setup, $A_t$ adheres to a truncated Gaussian distribution, $A_t \sim \mathcal{N}(A_{t-1}, \sigma_t^2 I, lb, ub)$, ensuring that the actions remain within plausible limits defined by physical and operational constraints of the vehicle. The term $\nabla_{S_t} \log q_t(S_t | S_0)$ represents the score function, crucial for the score matching approach. This gradient, which needs to be estimated via a neural network $s_\varphi(x_t, t)$, guides the correction of the forward model by quantifying how the probability density function of the state transitions should be adjusted to better fit the observed data.

This refined modeling through score matching not only enhances the accuracy of state prediction in backward time but also improves the robustness and adaptability of the navigation system under diverse and challenging driving conditions. The objective function can be written as:

$$\min_\theta \mathbb{E}_{t \sim u(0,T)} \mathbb{E}_{x_0 \sim q_0(x_0)} \mathbb{E}_{A_t \sim \mathcal{N}(A_{t-1}, \sigma_t^2 I, lb, ub)} \left\| s_\varphi(x_t, t) \right. \quad (17)$$
$$\left. - \nabla_{S_t} \log q_t(S_t) \right\|_2^2$$

where

$$S_t = \gamma_t S_0 + \sigma_t A \quad (18)$$

Since $A_t$ follows truncated normal distribution, the following equation holds when $A_t$ is within the boundaries:

$$\nabla_{S_t} \log q_t(S_t | S_0) = -\nabla_{S_t} \frac{(S_t - \gamma_t S_0)^2}{2\sigma_t^2} \quad (19)$$
$$= -\frac{S_t - \gamma_t S_0}{\sigma_t^2}$$
$$= -\frac{\gamma_t S_0 + \sigma_t A - \gamma_t S_0}{\sigma_t^2}$$
$$= -\frac{A}{\sigma_t}$$

Therefore

$$s_\varphi(x_t, t) \coloneqq -\frac{A_\theta(x_t, t)}{\sigma_t}, \quad (20)$$

For the timestep $t$ approaching a large number, we can use the reparameterization trick because $s_\varphi(x_t, t)$ would also follow Gaussian distribution based on the central limit theory. The objective of estimating the score function can be simplified as:

$$\min_\theta \mathbb{E}_{t \sim u(0,T)} \mathbb{E}_{x_0 \sim q_0(x_0)} \mathbb{E}_{A_t \sim \mathcal{N}(A_{t-1}, \sigma_t^2 I, lb, ub)} \left\| A_t \right. \quad (21)$$
$$\left. - A_\varphi(x_t, t) \right\|_2^2$$

However, the total timestep in the proposed forward process would have a limited length and the reparameterization trick is not applicable. In this case, a rolling back method is introduced to learn $\nabla_{S_t} \log q_t(S_t | S_{t+dt})$.

As illustrated in Fig. 5, the process demonstrates how a specific segment of state-action transitions from $S_{t-1} \xrightarrow{a_{t-1}} S_t$ in the forward phase is mirrored. In this reversal, the current state $S_t$ becomes $S_t'$, following the method outlined in Equation (7). A neural network model then predicts the probability $P(a_{t-1}' | S_t', \theta_t', l_t')$ using the reversed state $S_t'$ in stage ① as is illustrated in Fig. 5. By employing the mean and variance derived from $P(a_{t-1}' | S_t', \theta_t', l_t')$, a reparametrized action $z_{t-1}'$ is selected. Subsequently, a physical model, referred to as the bicycle model, is used to estimate the preceding state $S_{t-1}'$ from $S_t'$ and $z_{t-1}'$ in stage ② as is shown in Fig. 5. The discrepancy between this estimated state $S_{t-1}'$ and the actual prior state, once reversed to $S_{t-1}$, is quantified using the mean squared error (MSE). This rollback technique is pivotal in ensuring that the actions predicted by the model effectively guide the vehicle closer to its initial parking spot, thereby reversing its trajectory in a manner that approximates the original state transitions.

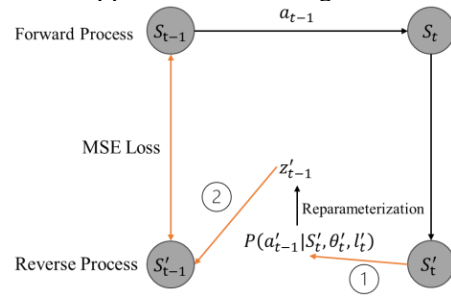

Fig. 5. Rolling back process.

This neural network model architecture, rooted in the principles of physics, utilizes a physics-informed variational autoencoder (VAE) approach to carry out two key stages of prediction, as is shown in Fig. 6. At its core, the model inputs the current reversed state $S'_t$. The initial phase involves an encoder, structured as a fully connected neural network, which predicts the likelihood of the previous action, $a'_{t-1}$ based on $S'_t$. This previous action is modeled to follow a Gaussian distribution, with the neural network providing the mean $\mu_{a'_{t-1}|x_t}$ and variance $\sigma^2_{a'_{t-1}|x_t}$ as outputs. Next, an action $z'_{t-1}$ is chosen using a technique known as reparameterization, which aids in drawing a sample from the predicted distribution. The next part of the model, which acts as a decoder, incorporates a physics-based framework to estimate the earlier state, $S'_{t-1}$, informed by the chosen action $z'_{t-1}$ and the current modified state $S'_t$. The process culminates with a comparison between the estimated previous state $S'_{t-1}$ and the actual previous state known from the data, referred to as $S''_{t-1}$. This final step verifies the accuracy of the model's predictions, ensuring that the chosen actions are effectively guiding the vehicle back toward a state that aligns with the known trajectory leading up to the original parking position.

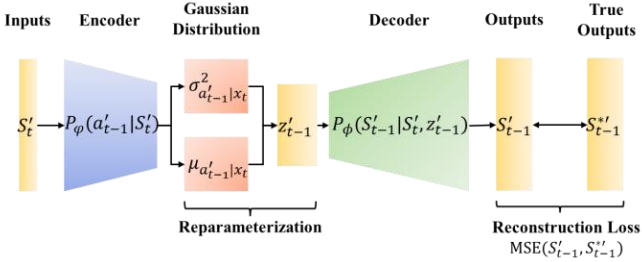

Fig. 6. Physics-informed VAE.

The loss function designed for the physics-informed VAE is threefold: Firstly, it accounts for the state prediction error, quantifying the difference between predicted states and true states. Secondly, it incorporates a regularization component for the variance $\sigma^2_{a'_{t-1}|x_t}$. This regularization ensures that the model does not overly concentrate on minimizing the prediction error linked to the mean $\mu_{a'_{t-1}|x_t}$, but also accurately gauges the level of uncertainty in the variance. The third term accounts for model uncertainty in predictions of the next state. It regulates the variance of state transition noise $\epsilon \sim \mathcal{N}(0, \Sigma)$. The structure of this loss function is intended to maintain a balance between precise state estimation and a reliable measure of prediction confidence.

$$\mathcal{L} = MSE(S'_{t-1}, S''_{t-1}) + \lambda_1 R_a\left(\sigma^2_{a'_{t-1}|x_t}\right) + \lambda_2 R_\epsilon(\Sigma) \quad (22)$$

where $\lambda_1$ and $\lambda_2$ serve as scaling factors that dictate the relative weight of the regularization term and the noise term in the overall loss function. By adjusting $\lambda_1$ and $\lambda_2$, one can control how much emphasis is placed on the regularization aspect, which governs the precision of the uncertainty captured by $R_a(\sigma^2_{a'_{t-1}|x_t})$ and $R_\epsilon(\Sigma)$, compared to the emphasis on minimizing the mean squared error (MSE) between the predicted state $S'_{t-1}$ and the true previous state $S''_{t-1}$. The value of $\lambda_1$ and $\lambda_2$ are chosen to balance the trade-off between accuracy of state prediction and reliability of the model's confidence in its predictions.

The function $R_a\left(\sigma^2_{a'_{t-1}|x_t}\right)$ defined in Equation (23) is a regularization term designed to refine the model's estimate of action variance at each timestep. It is expressed as the negative average over all timesteps $T$ of the logarithm of the variance $\sigma_i$ adjusted by a small constant $\epsilon$ to maintain numerical stability. Specifically, $\varepsilon$ prevents the logarithm from diverging to negative infinity in cases where the variance $\sigma_i$ approaches zero.

$$R_a\left(\sigma^2_{a'_{t-1}|x_t}\right) = -\frac{1}{T}\sum_{t=1}^{T}[log(\sigma_i + \varepsilon) + log(1 - \sigma_i - \varepsilon)] \quad (23)$$

This regularization term comprises two components: the log of $\sigma_i$ plus $\varepsilon$, and the log of $1 - \sigma_i - \varepsilon$, which together encourage the model not to be overly confident (by avoiding too small variance) or overly uncertain (by avoiding too large variance) about its action predictions. The balance achieved by this term is crucial for a model that needs to have a reasonable level of uncertainty to be robust yet confident enough to make accurate predictions. The noise loss term $R_\epsilon(\Sigma)$ defined in Equation (24) measures how likely the true next state is, given the model's predictions, scaled by the model's own uncertainty about its predictions (expressed through $\Sigma$).

$$R_\epsilon(\Sigma) = \frac{MSE(S'_{t-1}, S''_{t-1})}{2 \cdot \Sigma^2} \quad (24)$$

The function is fundamentally derived from the log-likelihood of a Gaussian distribution, which is a common approach in statistical modeling to handle errors or noise that follows a normal distribution. the log-likelihood of observing the next state $S'_{t-1}$ from a Gaussian distribution with the mean (true next state) $S''_{t-1}$ and variance $\Sigma^2$ is given by Equation (25):

$$log(p(S'_{t-1}|S''_{t-1}, \Sigma^2)) = -\frac{\left(S'_{t-1} - S''_{t-1}\right)^2}{2 \cdot \Sigma^2} - log(\Sigma\sqrt{2\pi}) \quad (25)$$

We can drop the constant term $log(\Sigma\sqrt{2\pi})$ for optimization since it does not affect the relative evaluations of the model parameters. This formulation leverages the properties of the Gaussian distribution to model uncertainty in state transitions and incorporate it into the loss function.

### C. Decisions on the move: Active Inference with the diffusion model

Active Inference (AIF) offers a framework for understanding and predicting the behavior of autonomous agents in dynamic and uncertain environments. This approach uses the principle of minimizing expected free energy and variational free energy to guide decision-making [19]. Variational free energy deals with the present (how well the agent's model predicts what it currently observes), while expected free energy is concerned with the future (choosing actions that minimize future surprise and maximize goal fulfillment) [20]. Together, these concepts help an agent continuously adapt and make informed decisions in a changing world, aiming for a coherent and accurate understanding of its environment and effective interaction with it.

Variational free energy is a concept derived from statistical physics but adapted in the realm of cognitive science to measure how well an agent's internal model predicts sensory inputs it observes. It is a discrepancy measure between what the agent expects to see and what it observes. Minimizing variational free energy means the agent is improving its model of the world to better predict incoming sensory data. In simpler terms, it's like an error signal that tells the agent how wrong its predictions were; by reducing this error, the agent's model becomes more accurate. Expected free energy, on the other hand, is more forward-looking. It measures not just the "fit" or accuracy of the agent's model against current observations but also considers the future states the agent might experience. Expected free energy considers the uncertainty or surprise that those future states could hold and how valuable they might be in terms of the agent's goals. By minimizing expected free energy, the agent doesn't just seek to reduce surprise in the present but also acts in ways that are expected to reduce surprise in the future while maximizing its goals [21].

Here, AIF is an approach that conceptualizes the way autonomous vehicles navigate and interact with the world. The automated parking is modeled as a finite horizon Markov decision process (MDP) [22]. The key to applying AIF is the balance between being true to vehicle's model of the parking environment while taking actions that are most likely to result in preferred outcomes, which are states that lead the vehicle back to its desired parking spot. The diffusion-based motion predictor operates as a generative model for a vehicle's interactions within its environment. This model, parameterized by $\varphi$, projects the likelihood of potential actions $p(a_n|S_n; \varphi)$, each action being one that could revert the vehicle to a position progressively nearer to its designated parking location. Complementing this predictive layer is the physical probabilistic bicycle model, symbolized as $f(\cdot)$, which serves to estimate the vehicle's subsequent state as a consequence of the selected actions. This physical model follows the dynamics encapsulated in the predefined Equations $(9) - (13)$, providing the trajectory that a vehicle would trace given a set of maneuvers. The interplay between the generative model and the physical model creates a cohesive framework for understanding and guiding a vehicle's movements towards its goal state.

$$S'_{n+1} = f(S_n, a_n) \qquad (26)$$

Incorporating the realities of dynamic environments into the predictive model, the likelihood of future states and actions can now be represented with a probability distribution that allows for environmental uncertainties such as slippery road conditions. This predictive distribution is described by:

$$Q(\vec{S'}, \vec{a}|S_n) := \qquad (27)$$
$$\prod_{\tau=n}^{N-1} \mathcal{N}(S_{\tau+1}; f(S_\tau, a_\tau), \Sigma) p(a_\tau|S_\tau; \varphi)$$

In Equation (27), $Q(\vec{S'}, \vec{a}|S_n)$ denotes the estimated future states and actions, assuming the agent is in state $S_n$. Instead of asserting that actions lead to a single specific state, the Gaussian distribution $\mathcal{N}$ introduces a scope of possible next states $S_{\tau+1}$, with $f(S_\tau, a_\tau)$ providing the mean or most likely next state and $\Sigma$ encapsulating the uncertainty in this transition. The term $p(a_\tau|S_\tau; \varphi)$ captures the conditional probability of an action $a_\tau$, given the current state $S_\tau$, and influenced by the model's parameters $\varphi$.

a. Refining the Predictive Model via Variational Free Energy

In the AIF framework, Variational Free Energy (VFE) serves as a measure of the divergence between the predicted and actual future states. In this work, VFE is used to quantify the discrepancy between the outcomes predicted by the diffusion model (DP model) and the observed true states. The VFE is formally represented by the equation:

$$VFE = D_{KL}[q(\varphi|S_n, a)||p(\varphi)] \qquad (28)$$
$$- E_q[\log p(\vec{S'}|S_n, a, \varphi)]$$

In equation (28), the first term is the KL divergence, which calculates the difference between the current belief about the model parameters $q(\varphi|S_n, a)$ and the prior beliefs $p(\varphi)$. The second term is the expected log probability of the observed states given the current state, actions, and model parameters, which serves to anchor the model's predictions to the actual observations. To optimize the diffusion model, VFE is minimized by continuously adjusting the model parameters, denoted as $\varphi$ using the data collected during operation. This optimization is typically performed using gradient descent on the physics-informed VAE, as is discussed in section III part b, where $\varphi_{new} = \varphi_{old} - \eta \nabla_\varphi VFE$, with $\eta$ is the learning rate.

b. Navigating Towards Goals with the Free Energy of the Future

In the exploration of preferred states within the AIF framework, a preference distribution $C_\beta$ is defined over the state space $\mathbb{S}$. This distribution is weighted by a parameter $\beta$, which is greater than zero, to prioritize states that the agent finds rewarding. Mathematically, preferred states are derived from the Boltzmann distribution expressed as in logarithmic form by:

$$-\log C_\beta(S) = --\beta R(S) - c(\beta), \forall S \qquad (29)$$
$$\in \mathbb{S}, \text{for some } c(\beta)$$
$$\in \mathbb{R} \text{ constant w.r.t s.}$$

The parameter $\beta$, known as the inverse temperature, quantifies the agent's motivation level: a higher $\beta$ corresponds to a stronger preference for states yielding higher rewards. Agents are therefore inclined to select states that maximize the reward function $R(S)$, thus maximizing $C_\beta(S)$ and minimizing $-\log C_\beta(S)$ for any given $\beta$ greater than zero. This foundational preference structure underpins the agent's decision-making process, steering it toward states that it deems preferable or beneficial in the context of its environment and objectives. The reward function $R(S)$ captures the criteria for these desired states, integrating goals such as reaching a parking spot $S_{goal}$, maintaining safety by avoiding collisions with surrounding vehicles $S_{n+1}^{v-}$ at time step $n+1$, and ensuring smoothness in the control actions. Mathematically, the reward function is characterized by:

$$R(S) = -\lambda_{goal} \cdot \|S'_{n+1} - S_{goal}\| + \lambda_{safety} \qquad (30)$$
$$\cdot \sum \|S'_{n+1} - S_{n+1}^{v-}\|$$
$$- \lambda_{smooth} \cdot \|a'_n - a_{n-1}\|$$

where $\lambda_{goal}$, $\lambda_{safety}$, and $\lambda_{smooth}$ are the weighting parameters that balance the importance of each aspect in the reward function. Extending the preference distribution $C_\beta$ over trajectories $\vec{S} := (S_1, S_2, \cdots, S_N) \in \mathbb{S}^N$, we apply the additive property of the reward function to evaluate entire paths:

$$-log\, C_\beta(\vec{S}) = -\beta R(\vec{S}) - c'(\beta) \qquad (31)$$
$$= -\sum_{\tau=1}^{N} \beta R(S_\tau) - c'(\beta), \forall \vec{S}$$
$$\in S^N$$

The inverse temperature parameter $\beta$ remains a measure of how strongly the agent prefers certain trajectories, favoring those that accumulate greater rewards. Through this framework, the active inference process not only seeks individual states but also entire trajectories that are aligned with the agent's preferences and the dynamics of the vehicle's environment.

Vehicles aim to minimize a quantity known as Expected Free Energy (EFE) in the framework of active inference, which guides them in making decisions that align with their preferences. This aims at balancing the exploration-exploitation trade-off by minimizing surprise (or uncertainty) and maximizing the likelihood of achieving preferred outcomes [20]. The general formula for EFE is:

$$G(s,a) = \mathbb{E}_{Q(s'|s,a)}[\log Q(s'|s,a) \qquad (32)$$
$$- \log P(o, s'|s,a)]$$

where $s$ represents the current state, $a$ represents the action taken by the agent, $s'$ represents the subsequent state resulting from action $a$, and $o$ is the observation or outcome associated with state $s'$. $Q(s'|s,a)$ is the approximate posterior or the agent's belief about the next state given the current state and action. $P(o,s'|s,a)$ is the generative model that links states and observations, providing the likelihood of observing $o$ in state $s'$ after taking action $a$. The term $\log Q(s'|s,a)$ refers to the entropy of the agent's beliefs, which measures uncertainty or surprise about the next state. The term $\log P(o,s'|s,a)$ captures the accuracy of the predictions under the model, quantifying how probable the outcomes are given the agent's model of the world. In practice, this formula guides agents to choose actions that are expected to provide the most informative (reducing uncertainty) and rewarding outcomes according to their internal model of the world.

Equation (33) can be approximated and split into an energy and an entropy or an accuracy and complexity term [20], which corresponds to the extrinsic and epistemic action terms in the EFE:

$$G(s,a) \qquad (33)$$
$$\approx -\mathbb{E}_{Q(s'|s,a)}[\log P(o, s'|s,a)]$$
$$+ \mathbb{E}_{Q(s'|s,a)} D_{KL}[Q(\overrightarrow{S'}|\vec{a}, S_n)|C_\beta(\vec{S})]$$

where the first term $\mathbb{E}_{Q(s'|s,a)}[\log P(o,s'|s,a)]$ is the extrinsic value. It quantifies the surprise or improbability of observing $o$ and the next state $s'$ given the current state $s$ and the current action $a$, thereby estimating how unexpected or unlikely these observations and transitions are under the current policy. The second term $\mathbb{E}_{Q(s'|s,a)} D_{KL}[Q(\overrightarrow{S'}|\vec{a}, S_n)|C_\beta(\vec{S})]$ is the intrinsic value, which incorporates the cost function $C_\beta(\vec{S})$ and quantifies how the distribution of predicted future states $Q(\overrightarrow{S'}|\vec{a}, S_n)$ diverges from a desired or preferred state distribution as encoded by $C_\beta(\vec{S})$. This divergence aims to penalize decisions leading to future states that are less preferred according to the cost function. Essentially, it encourages the selection of actions that not only minimize surprise but also align future states closely with those that are considered preferable or beneficial.

Given a MDP process, the EFE for a sequence of actions, denoted as $G(\vec{a}|s_t)$, can be expressed as an aggregate of individual free energies at each time step:

$$G(\vec{a}|s_t) \approx \sum_{\tau=n+1}^{N} G(a_\tau|s_\tau) \qquad (34)$$

This simplification allows the agent to plan by evaluating each future time point separately, significantly streamlining the planning process without the need for an exhaustive evaluation of all possible future trajectories. It is an efficient method to guide the agent toward preferred states while considering the inherent uncertainties and computational constraints.

### D. Comprehensive Workflow of the Diffusion-Based Active Inference Framework

The following algorithm outlines the entire workflow of the Diffusion-Based Active Inference Framework (AIF). It is structured into three interlinked phases: Forward Diffusion Process, Reversed Diffusion Process, and Active Inference Control. Using simulation for training, vehicles exit parking spots under a variety of initial conditions, performing random maneuvers such as steering and throttle adjustments. The process continues until the vehicle either collides or reaches the boundary of the parking area. The reversed Diffusion process utilizes the data generated in the forward Diffusion process and employs a physics informed Variational Autoencoder (VAE) model to reverse the sequence of the collected data. The model predicts previous states and actions from current states, enabling the vehicle to reverse-engineer its movements. The reverse model is refined through VFE when it fails to properly predict. The refined reverse model and AIF then decide the most probable action(s) that minimize expected free energy. the vehicle toward its intended destination, dynamically adapting to new data and making necessary course corrections.

| **Algorithm 1: Algorithm for Diffusion-Based Active Inference Framework** |
|---|
| ***DP-Forward:*** *Data Collection* |
|    ***Input:*** *Number of trials, vehicle dynamics.* |
|    ***Output:*** *Vehicle trajectory dataset.* |
| **1**   *For each trial:* |
| **2**      *Initialize parking lot with vehicles at random spots.* |
| **3**      ***Simulate vehicle movement until a collision or boundary is reached:*** |
| **4**         *Apply random actions (throttle, steering) and introduce Gaussian noise to model uncertainty.* |
| **5**         *Record state transitions and actions until the end condition is met.* |
| ***DP-Reverse:*** *Model Training* |
|    ***Input:*** *Collected vehicle trajectories.* |
|    ***Output:*** *Trained physics-informed VAE model for* |

*motion prediction.*

6  *Reverse the state-action sequences from the trajectory dataset.*

7  *Initialize the VAE model.*

8  **For each training epoch:**

9  *Predict action distributions to reverse states.*

10  *Sample actions using reparameterization.*

11  *Predict previous states using the probabilistic bicycle model.*

12  *Compute loss and update model parameters.*

**Active Inference Control:** *Implementation and Execution*

**Input:** *Pretrained VAE model, vehicle's current state, and destination.*

**Output:** *Vehicle actions (throttle, steering).*

13  *Continuously predict and apply actions that minimize EFE until the vehicle reaches the desired spot.*

14  **Monitor state prediction error:**

15  *If error exceeds a threshold, fine-tune the model with recent state-action data.*

The algorithm is inherently cyclic and adaptive, featuring feedback mechanisms within the Active Inference Control phase that can trigger additional data collection and model retraining as needed. This adaptive cycle ensures that the navigation system continuously evolves, enhancing its capacity to handle increasingly complex environments and improving accuracy over time.

## IV. SIMULATION AND VALIDATION RESULTS

In this section, we delve into the validation results from simulations conducted within a custom-designed parking environment created using the 'highway-env' simulation package. The purpose of these simulations is to assess the real-world applicability and robustness of the proposed Framework. The validation is divided into three segments. The initial part describes the simulation environment set up within 'highway-env'. This setup provides the foundational context for subsequent testing and analysis. In the second segment, a detailed analysis of the diffusion model's performance is presented. Finally, we examine the simulation outcomes where the model's practical efficacy is showcased through its application in controlling vehicles. These simulations underscore the model's potential contributions to autonomous navigation under uncertainty.

### A. Simulation Setup

a. Setup for training the diffusion motion predictor

For the motion prediction model, the highway-env simulation package was tailored to create two specific parking scenarios, shown in figure 7, with the aim of training and validating a diffusion-based motion predictor. In the simulation setup, green vehicles are designated for autonomous control, whereas the yellow vehicles remain stationary. The parking area is outlined by yellow lines, denoting its boundaries. The first scenario is structured with six parking spots and four cars within the simulated environment shown in figure 7(a). Two of these vehicles are designated as controllable, while the remaining pair are static and parked. In the second, more complex scenario, the environment is expanded to include ten parking spots and six cars, with half of the vehicles being under our control, as is shown in figure 7(b).

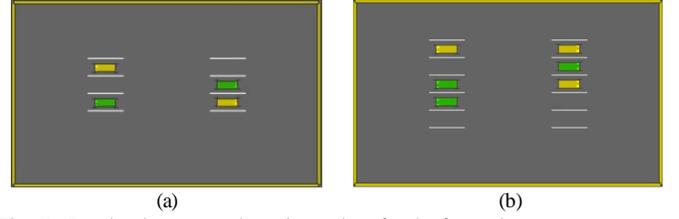

(a)        (b)

Fig. 7. Randomly generated starting points for the forward process.

For training phase, the starting positions of the vehicles are randomized in different parking spots in each trial, ensuring a diverse range of initial conditions for model training.

b. Customizing the Simulation for AIF-Controlled Driving

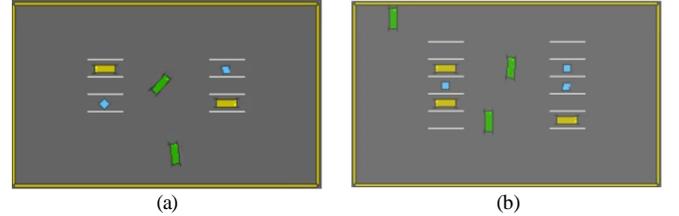

(a)        (b)

Fig. 8. Randomly generated starting points for the reverse process.

For the application phase, the parking environment is again customized for scenarios with four and six cars as is shown in Fig. 8. Parked cars are placed in predetermined spots, while the controllable vehicles are placed at random locations within the parking area, each with an assigned destination spot. The controllable cars, equipped with initial velocities and orientations, utilize AIF for navigation, driving towards their designated parking spots while avoiding collisions. This two-tiered simulation approach serves a dual purpose: training the model to understand vehicle dynamics and control strategies in a constrained environment and validating the model's capability to navigate complex scenarios with multiple agents. The results from these simulations are expected to provide insightful data on the potential of AIF in the field of autonomous vehicle navigation, particularly in unstructured environments where traditional driving guidelines may be absent.

### B. Assessing the performance of the diffusion motion predictor

Fig. 9 illustrates the loss plot of the diffusion motion predictor throughout its training and validation phases. Initially, a precipitous decline in the training loss is observed, indicative of the model's rapid acclimatization to the structure within the training data. Concurrently, the validation loss mirrors the downward trend of the training loss, suggesting a consistent learning pattern that generalizes beyond the training set. As the epochs advance, both losses stabilize and exhibit minimal variance, which implies that the model has potentially reached its learning capacity given the current architecture and dataset. The absence of a significant gap between the training and validation losses towards the end of the training suggests that the model is not overfitting and is well-calibrated to the complexity of the data it aims to model.

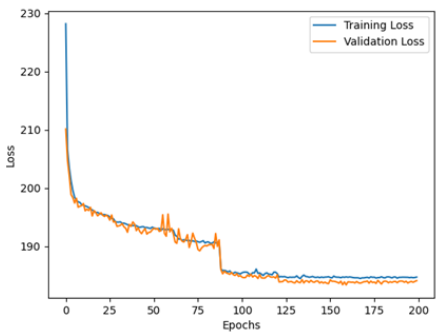

Fig. 9. Loss plot for training the reverse process.

Table I presents some prediction examples that illustrates the diffusion-based motion predictor's performance. The table compares the positions of the controlled vehicles after applying the reparametrized action selected from the predicted action distribution against their actual positions. These examples indicate the model's adeptness in tracking and predicting vehicle motion with a high degree of accuracy, as reflected by the minor discrepancies between predicted and true states.

TABLE I
PREDICTION EXAMPLES OF THE DIFFUSION-BASED MOTION PREDICTOR

| # | Next State | $x$ | $y$ | $v_x$ | $v_y$ | $h$ |
|---|---|---|---|---|---|---|
| 1 | True next state | 3.96 | -19.46 | -0.86 | 1.59 | -1.08 |
|   | Predicted next state | 3.92 | -19.55 | -0.73 | 1.65 | -1.15 |
| 2 | True next state | -11.36 | -8.30 | 9.80 | 9.73 | -2.36 |
|   | Predicted next state | -11.61 | -7.58 | 8.67 | 10.76 | -2.25 |
| 3 | True next state | 6.16 | 14.55 | -0.08 | -1.41 | 1.51 |
|   | Predicted next state | 6.32 | 14.47 | -0.11 | -1.78 | 1.51 |

Table II enumerates more key metrics used to evaluate the diffusion-based motion predictor. The Mean Squared Error (MSE) for the model's predictions, which evaluates the average squared difference across all elements in the vehicle state ($x$, $y$, $v_x$, $v_y$, and $h$), is 0.2296. indicating the average squared difference between the predicted and actual next states. Furthermore, the model demonstrates a probabilistic confidence in its predictions, with the true next state falling within one standard deviation (1 sigma) of the predicted distribution 43.05% of the time, within two standard deviations (2 sigma) 68.22% of the time, and within three standard deviations (3 sigma) 86.27% of the time. These statistics not only affirm the model's predictive strength but also suggest a well-calibrated understanding of the uncertainty inherent in vehicle movements.

TABLE II
PERFORMANCE METRICS FOR THE DIFFUSION-BASED MOTION PREDICTOR

| Metric Description | Value |
|---|---|
| MSE between reparametrized next state and true next state | 0.2296 |
| True next state within 1 $\delta$ of predicted next state distribution | 43.05% |
| True next state within 2 $\delta$ of predicted next state distribution | 68.22% |
| True next state within 3 $\delta$ of predicted next state distribution | 86.27% |

### C. Active Inference Framework: Guiding Autonomous Vehicles to Precision Parking

The trajectory plots shown in Fig. 10 illustrates the behaviors of the vehicles under the guidance of AIF. Fig. 10 (a) and (b) shows the trajectories for two controlled vehicles, (c) and (d) shows the cases with three controlled vehicles. They depict the routes taken from the vehicles' starting points to their parking spots, highlighting the adaptive maneuvers made to avoid obstacles and achieve their parking goals. Through these plots, we can demonstrate AIF's capacity for spatial reasoning and its application in complex navigation tasks, validating its use in autonomous parking systems.

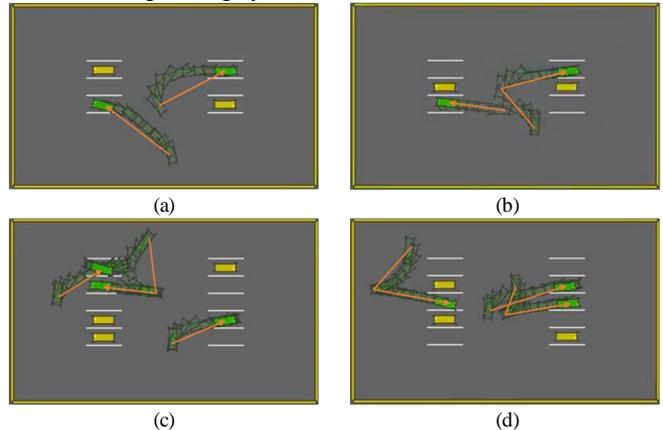

(a)  (b)

(c)  (d)

Fig. 10. Vehicle trajectories controlled by diffusion-based AIF.

## V. CONCLUSION AND FUTURE WORK

In this paper, we have explored a novel approach to guiding autonomous vehicles in scenarios where the usual road markings are absent. This exploration was grounded in the development of a diffusion-based motion predictor, implemented within an Active Inference Framework (AIF), and tested within a specially designed parking lot simulation. Our goal was to closely mimic the challenges vehicles face on unmarked roads, using the parking lot as a stand-in for such environments. We started by transforming a traditional road scenario into a parking lot setup, a crucial step in creating a realistic yet controlled environment for our simulations. This environment, characterized by its lack of lane markings, required vehicles to navigate from their starting points to designated parking spots while avoiding collisions. This scenario was meticulously designed to transition vehicles from a disordered state, where their positions and velocities were randomized, to an orderly state, mirroring the structured outcome of successful parking. The introduction of the diffusion-based motion predictor was a pivotal part of our exploration. This tool, developed through a nuanced understanding of diffusion model methodologies, was adept at forecasting the future movements of vehicles within the parking lot. By predicting a range of possible actions for each vehicle and selecting the optimal path based on expected free energy, the model demonstrated its ability to effectively navigate vehicles to their intended destinations.

Our simulations offered concrete evidence of the model's effectiveness. Through a series of tests in environments with varying degrees of complexity, from four to six cars navigating towards six to ten parking spots, we showcased the model's robust capability to ensure safe and efficient vehicle parking. These tests not only affirmed the model's practical applicability but also its potential to significantly improve traffic safety and vehicle navigation in real-world scenarios devoid of lane markings. Looking forward, we aim to extend the scope of our research to encompass larger and more complex driving environments. The next steps involve refining the model to enhance its predictive accuracy and adaptability to the unpredictability inherent in real-world driving conditions. The scalability of the proposed system will also be tested in urban driving environments, pushing the boundaries of what's currently achievable in autonomous vehicle navigation.

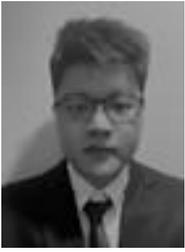

**Yufei Huang** received a B.Eng. degree in Automation from Xi'an Jiaotong University in 2016. Also, he received an M.S. degree in Systems Engineering from the University of Maryland, College Park in 2018. He is currently a Ph.D. student at Rutgers, the State University of New Jersey, studying Industrial and Systems Engineering, and a Research Assistant at the Center for Advanced Infrastructure and Transportation (CAIT). His research interests are in multi-agent systems, autonomous systems, robotics, and reinforcement learning.

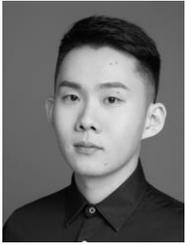

**Yulin Li** received his B.S. degree in industrial and systems engineering from Rutgers University in 2019 and an M.S. degree in systems engineering from Cornell University in 2021. He is currently working towards a Ph.D. degree in industrial engineering at Rutgers University. His research interests include generative models and the use of active inference for decision-making in dynamic environments.

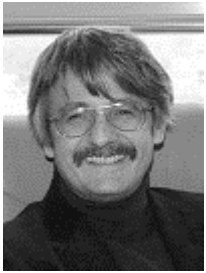

**Andrea Matta** is Full Professor at Politecnico di Milano, where he currently teaches integrated manufacturing systems and manufacturing processes. His research area includes analysis, design, and management of manufacturing and healthcare systems. He is Editor in Chief of the Flexible Services and Manufacturing Journal.

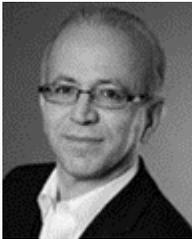

**Mohsen Jafari** (M'97) received a Ph.D. degree from Syracuse University in 1985. He has directed or co-directed a total of over 23 million U.S. dollars in funding from various government agencies, including the National Science Foundation, the Department of Energy, the Office of Naval Research, the Defense Logistics Agency, the NJ Department of Transportation, FHWA, and industry in automation, system optimization, data modeling, information systems, and cyber risk analysis. He actively collaborates with universities and research institutes abroad. He has also been a Consultant to several Fortune 500 companies as well as local and state government agencies. He is currently a Professor and the Chair of Industrial & Systems Engineering at Rutgers University-New Brunswick. His research applications extend to manufacturing, transportation, healthcare, and energy systems. He is a member of the IIE. He received the IEEE Excellence Award in service and research.